\pdfoutput=1
\documentclass[11pt]{article}

\usepackage[]{EMNLP2022}

\usepackage{times}
\usepackage{latexsym}

\usepackage{graphicx}
\usepackage{amsthm,amsmath,amssymb}
\usepackage{mathrsfs}
\usepackage{amssymb}
\usepackage{booktabs}
\usepackage{CJKutf8}

\usepackage[T1]{fontenc}
\usepackage[utf8]{inputenc}

\usepackage{microtype}

\usepackage{inconsolata}

\title{Third-Party Aligner for Neural Word Alignments}

\usepackage{hyperref}

\makeatletter
\newcommand{\printfnsymbol}[1]{%
  \textsuperscript{\@fnsymbol{#1}}%
}
\makeatother

\author{
	{Jinpeng Zhang\textsuperscript{1}\thanks{\ \ \ Equal Contribution. }\ , Chuanqi Dong\textsuperscript{1}\textit{\footnotemark[1]}\ , Xiangyu Duan\textsuperscript{1}\thanks{\ \ \  Corresponding Author. }\ , Yuqi Zhang\textsuperscript{2}\ ,\ Min Zhang\textsuperscript{1} } 
	\vspace{2.0mm}\\
	\fontsize{12}{10}\selectfont
	\,\textsuperscript{\rm 1}  Institute of Aritificial Intelligence, School of Computer Science and Technology, \\
	\fontsize{12}{10}\selectfont Soochow University  \\
            \fontsize{12}{10}\selectfont  \textsuperscript{\rm 2} Alibaba DAMO Academy \\
            \fontsize{10}{10}\selectfont \{jpzhang1,cqdong\}@stu.suda.edu.cn; xiangyuduan@suda.edu.cn;\\
	\fontsize{10}{10}\selectfont chenwei.zyq@alibaba-inc.com; minzhang@suda.edu.cn\\	} 
\begin{document}
\maketitle
\begin{abstract}
Word alignment is to find translationally equivalent words between source and target sentences. Previous work has demonstrated that self-training can achieve competitive word alignment results. In this paper, we propose to use \emph{word alignments generated by a third-party word aligner} to supervise the neural word alignment training. Specifically, source word and target word of each word pair aligned by the third-party aligner are trained to be close neighbors to each other in the contextualized embedding space when fine-tuning a pre-trained cross-lingual language model. Experiments on the benchmarks of various language pairs show that our approach can surprisingly do self-correction over the third-party supervision by finding more accurate word alignments and deleting wrong word alignments, leading to better performance than various third-party word aligners, including the currently best one. When we integrate all supervisions from various third-party aligners, we achieve state-of-the-art word alignment performances, with averagely more than two points lower alignment error rates than the best third-party aligner.We released our code at \url{https://github.com/sdongchuanqi/Third-Party-Supervised-Aligner}.
\end{abstract}

\section{Introduction}

Word alignment is to find the correspondence between source side and target side words in a sentence pair \cite{brown1993}. It is widely applied in a variety of natural language processing (NLP) tasks, including learning translation lexicons \cite{Ammar2016,cao2019multilingual}, cross-lingual transfer \cite{yarowsky2001,pado2009,tiedemann2014,agic2016multilingual,mayhew2017cheap,nicolai2019learning}, and semantic parsing \cite{herzig2018}. In particular, word alignment plays a key role in many neural machine translation (NMT) related methods, such as imposing lexical constraints in the decoding process \cite{arthur2016,hasler2018}, improving automatic post-editing \cite{pal2017}, guiding learned attention \cite{liu2016neural}, and automatic analysis or evaluation of NMT models \cite{tu2016,bau2018identifying,stanovsky2019,neubig2019compare,wang2020inference}.

\begin{figure}[tb]
    \centering
    \includegraphics[width=7.5cm]{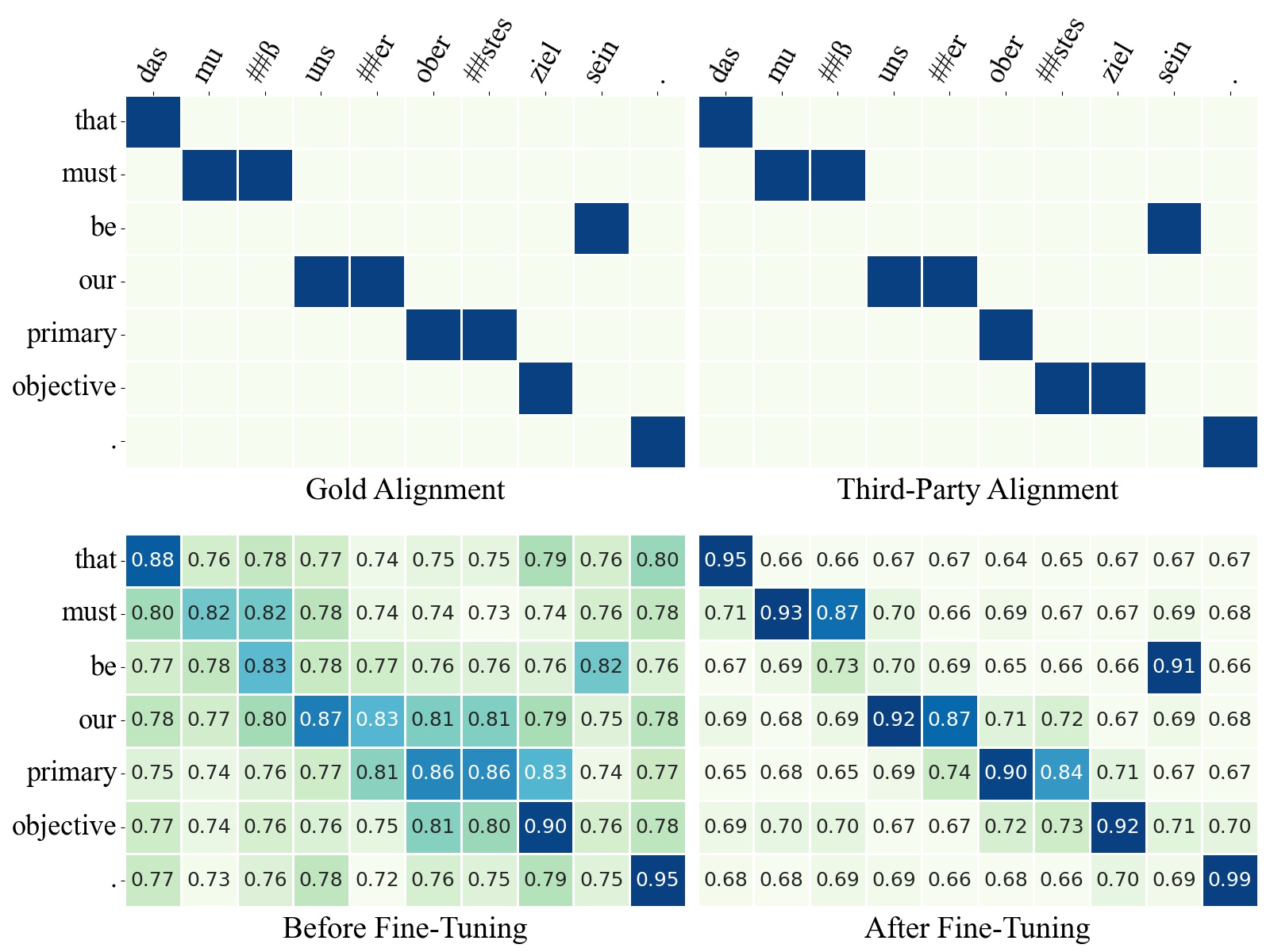}
    \caption{Gold and third-party word alignments, and cosine similarities between contextualized embeddings of subwords in a parallel sentence pair before and after ﬁne-tuning with third-party supervision.}
    \label{fig:illustration}
\end{figure}

Word alignment is usually inferred by GIZA++ \cite{och2003systematic} or FastAlign \cite{dyer2013simple}, which are based on the statistical IBM word alignment models \cite{brown1993}. Recently, neural methods are applied for inferring the word alignment. They use NMT-based framework to induce alignments through using attention weights or feature importance measures, and surpass the  statistical word aligners such as GIZA++ on a variety of language pairs \cite{li2019word,garg2019jointly,zenkel2019adding,zenkel2020end,chen2020accurate,song2020alignment,song2020towards,chen2021mask}. 

Inspired by the success of the large-scale cross-lingual language model (CLM) pre-training \cite{devlin2019bert,conneau2019cross,conneau2020unsupervised}, the pre-trained contextualized word embeddings are also explored for the word alignment task by either extracting alignments based on the pre-trained contextualized embeddings \cite{sabet2020simalign} or fine-tuning the pre-trained CLMs by self-training to get new contextualized embeddings appropriate for extracting word alignments \cite{dou2021word}. Based on careful design of self-training objectives, the fine-tuning approach achieves competitive word alignment results \cite{dou2021word}.

In this paper, we use simple supervision instead of the self-training to fine-tune the pre-trained CLMs. The simple supervision is derived from a third-party word aligner. Given a parallel corpus, the third-party word aligner predicts the word alignments over the corpus, which are used as the supervision signal for the fine-tuning. In particular, for each aligned word pair of a parallel sentence pair, the contextualized embeddings of the source and target words are trained to have high cosine similarity to each other in the embedding space.

As illustrated by Figure \ref{fig:illustration}, the cosine similarities between the source and target words of the correct word alignments are not quite high before the fine-tuning. The third-party word aligner can provide some correct word alignments (e.g. ``that'' ``must'' ``be'' associated alignments) along with wrong ones (e.g. ``primary'' ``objective'' associated alignments) as the supervision. Although the supervision is not perfect, it is still helpful for driving the contextualized embeddings of the source and target words of a correct word alignment closer in the embedding space after the fine-tuning. Surprisingly, with imperfect third-party supervision in fine-tuning, the heat map of the cosine similarities exhibits clearer split between the correct and wrong word alignments than not fine-tuning. Wrong alignments of the third-party aligner are rectified after fine-tuning (e.g. ``primary'' ``objective'' associated alignments), and the incorrect alignment before fine-tuning (e.g. ``be''associated alignment) is also rectified after fine-tuning.

We perform experiments on word alignment benchmarks of five different language pairs. The results show that the proposed third-party supervising approach outperforms all third-party word aligners. When we integrate all supervisions from various third-party word aligners, we achieve state-of-the-art performances across all benchmarks, with an average word error rate two points lower than that of the best third-party word aligner.

\begin{figure}[tb]
    \centering
    \includegraphics[width=8.5cm]{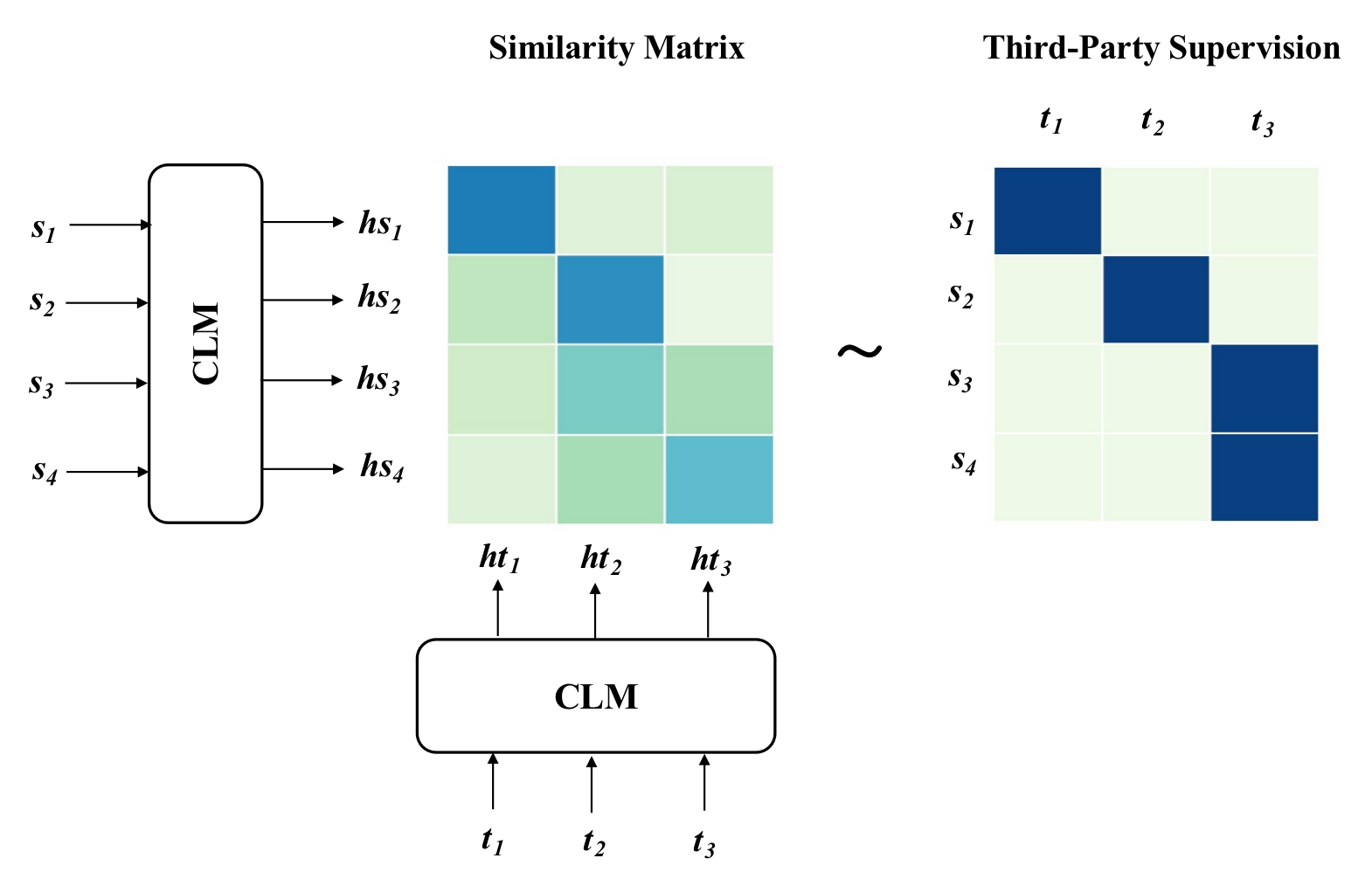}
    \caption{The framework of the fine-tuning with the third-party supervision.}
    \label{fig:framework}
\end{figure}

\section{Approach}

Formally, the word alignment task can be defined as finding a set of word pairs in the sentence pair $\langle${\bf{s}, \bf{t}}$\rangle$, where {\bf{s}} denotes the source sentence ``$s_1, ..., s_n$'', and {\bf{t}} denotes the corresponding target sentence ``$t_1, ..., t_m$'' parallel to {\bf{s}}. The set of the word pairs is: 

\begin{equation}
A=\{\langle s_i, t_j \rangle | s_i \in {\bf{s}}, t_j \in {\bf{t}} \}.   \nonumber 
\end{equation}

\noindent In each word pair $\langle s_i, t_j \rangle$, $s_i$ and $t_j$ are translationally equivalent to each other within the context of the sentence pair.

In the following, we will describe how we obtain the word alignments by fine-tuning the pre-trained CLMs. Different to previous work that fine-tunes by self-training \cite{dou2021word}, we supervise the fine-tuning process with third-party word alignments. 

\subsection{Third-Party Supervision}

The large-scale CLM pre-training has gained impressive performances across various NLP tasks \cite{libovicky2019language,hu2020xtreme}. As the outcome of the pre-trained CLMs, the contextualized word embeddings can represent words in semantic context across different languages. By further fine-tuning the CLMs, the contextualized embeddings of the source and target words of a word alignment in the embedding space can become closer, which makes it easier for identifying word alignments according to the simple geometry of the embedding space for each pair of parallel sentences.

We propose to fine-tune the pre-trained CLMs with supervision from a third-party word aligner. Figure \ref{fig:framework} shows the overall fine-tuning framework. For a source sentence ``$s_1, s_2, s_3, s_4$'' and its corresponding target sentence ``$t_1, t_2, t_3$'', we stack CLM over them to obtain the contextualized word embeddings {\bf{hs}} =``$hs_1, hs_2, hs_3, hs_4$'' and {\bf{ht}} =``$ht_1, ht_2, ht_3$'' for the source and target sides, respectively. Since CLM models sentences of different languages in the same contextualized embedding space, it is easy to constitute a similarity matrix by directly computing the cosine similarities between {\bf{hs}} and {\bf{ht}}. The similarity matrix is: 

\begin{equation}
M= {\bf{hs}} \times {\bf{ht}}^{\rm T}     \nonumber 
\end{equation}

In the matrix, word pairs with higher similarities are deemed as word alignments. Let $A'$ denotes the word alignments generated by the third-party word aligner. CLM is fine-tuned with the supervision of $A'$ so that $M$ is consistent with $A'$. Although the third-party supervision $A'$ is not perfect, we observe that the fine-tuning can proceed with self-correction of imperfect $A'$ in the experiments.

The supervision is bidirectional: 

\begin{align}
P&_{s2t}(i,j) =  \frac{e^{M_{i,j}}}{\sum_{j=1}^n e^{M_{i,j}}} \hfill  \notag \\
P&_{t2s}(i,j) =  \frac{e^{M_{i,j}}}{\sum_{i=1}^m e^{M_{i,j}}}   \hfill  \notag  \\
\mathcal{L} & = \frac{1}{m} \sum_{i=1}^{m} {\sum_{ \substack{j \\ s.t.\ \langle s_i,t_ j \rangle  \in A'} } P_{s2t}(i,j) } \hfill \notag \\
         & + \frac{1}{n} \sum_{j=1}^{n} {\sum_{ \substack{i \\ s.t. \ \langle s_i,t_ j \rangle  \in A'} } P_{t2s}(i,j) } \hfill  \label{equ1}
\end{align}

\noindent where $P_{s2t}$ denotes the probability of source-to-target alignment between $s_i$ and $t_j$, which is computed by softmax over the $i$th row of $M$. Correspondingly, $P_{t2s}$ denotes the probability of target-to-source alignment between $t_j$ and $s_i$, which is computed by softmax over the $j$th column of $M$\footnote{Since the experimental result difference between softmax and $\alpha$-entmax \cite{peters2019sparse} is marginal, we adopt softmax for simplicity. }. $m$ and $n$ denote the lengths of the source and target sentences, respectively. We aim to maximize $\mathcal{L}$, which sums the bidirectional probabilities subject to $A'$ supervision. 

Through the above training objective, CLM is fine-tuned to generate the contextualized embeddings suitable for building the similarity matrices to extract word alignments.

\subsection{Word Alignment Prediction}

Given a new pair of parallel sentences in the test set, we can predict its word alignments based on the CLM fine-tuned on the parallel training corpus. In particular, for the sentence pair, the source-to-target probability matrix $M_{s2t}$ which consists of probabilities of $P_{s2t}$, and the target-to-source probability matrix $M_{t2s}$ which consists of probabilities of $P_{t2s}$, are computed using the fine-tuned CLM at first, then the set of word alignments $A$ can be deduced according to the intersection of the two matrices:

\begin{equation}
A = \{\langle s_i, t_j \rangle | P_{s2t}(i,j) > c \ \ \& \ \  P_{t2s}(i,j) > c  \}     \nonumber 
\end{equation}

\noindent where $c$ is a threshold. Only the word pairs whose source-to-target alignment probability and target-to-source alignment probability are both greater than $c$ are deemed as the predicted word alignments. 

\subsection{Integrating various Third-Party Supervisions}

Different third-party word aligners exhibit different behaviors in the word alignment results. We integrate the word alignments produced by various aligners into one set of supervisions for the fine-tuning process to test if they can be combined to improve the performance further. At first, we group all third-party aligners' output alignments into one union. Then we utilize the union in two category of methods: filtering and weighting. 

The filtering method abandons word alignments in the union which have low consistency between various aligners, and only keep the alignments that majority of the aligners consent to. The remaining word alignments are used to supervise the fine-tuning process. Since different aligners get different performances, we assign credit to each aligner by using its performance on the development set (i.e., 
negative alignment error rate of the development set), then we normalize the credits of all aligners by softmax. Consequently, each word alignment $\langle s_i, t_j \rangle$ in the union has a total credit:

$$Credit_{total}(i,j) = \sum_{\substack{ k=1 \\ s.t.\  \langle s_i, t_j \rangle \in A'_{k} } }^{K} Credit_{k}(i,j) $$

\noindent where $A'_{k}$ denotes the set of word alignments of the $k$th third-party word aligner, $K$ denotes the number of the third-party word aligners, and $Credit_{k}$ is the credit of the $k$th aligner after softmax. $Credit_{total}$ represents the degree of agreement between various aligners. Only word alignments whose $Credit_{total}$ are greater than a threshold $f$ are kept for the subsequent fine-tuning.

Different to the filtering method, the weighting method considers all word alignments in the union, though it put weights over them in the fine-tuning. $Credit_{total}$ used in the filtering method is also adopted in the weighting method:

$$ w_{i,j} = \frac{1}{1+e^{-\lambda (Credit_{total}(i,j) - f)}} $$

\noindent where $w_{i,j}$ is the weight of the word pair $\langle s_i, t_j \rangle$. When $Credit_{total}$ exceeds the threshold $f$, the weight tends to 1, otherwise it tends to 0. $\lambda$ is the hyper-parameter that controls the effect of the supervision integration. $w_{i,j}$ is inserted into the fine-tuning objective $\mathcal{L}$ in euqation (\ref{equ1}) by simply replacing $P_{s2t}(i,j)$ with $w_{i,j} P_{s2t}(i,j)$, and replacing $P_{t2s}(i,j)$ with $w_{i,j} P_{t2s}(i,j)$.

\subsection{Handling Subwords}

Subwords \cite{sennrich2016neural,wu2016google} are widely used in pre-training CLMs. The fine-tuning process is conducted on the contextualized embeddings of the subwords. So we run all third-party word aligners at the subword level to get subword alignments, which are used for supervising the fine-tuning. During testing, we get the subword alignments for the test set sentence pairs at first, then convert the subword alignments to the word alignments by following previous work \cite{sabet2020simalign,zenkel2020end}, which consider two words to be aligned if any of their subwords are aligned.

\section{Experiments}

We test the proposed third-party supervised fine-tuning approach on word alignment tasks of five language pairs: Chinese-English (Zh-En), German-English (De-En),  English-French (En-Fr), Romanian-English (Ro-En) and Japanese-English (Ja-En). 

\subsection{Datasets}
We use the benchmark datasets of the five language pairs. They are utilized in two ways. For all third-party aligners, whole training corpus for each language pair is used by each third-party aligner. For our approach, only a fraction of the whole training corpus for each language pair is used in the fine-tuning phase.

Regarding the datasets for all third-party aligners, the configuration is the same to previous works. The Zh-En training-set is from the LDC corpus which consists of 1.2M sentence pairs, and the test and development sets are obtained from the TsinghuaAligner website \footnote{\url{http://nlp.csai.tsinghua.edu.cn/~ly/systems/TsinghuaAligner/TsinghuaAligner.html}} \cite{liu2005log}. For the De-En, En-Fr, Ro-En datasets, we follow the experimental setup of previous work\cite{zenkel2019adding,zenkel2020end} and use their pre-processing scripts \cite{zenkel2019adding}\footnote{\url{https://github.com/lilt/alignment-scripts}} to get the training and test sets. The Ja-En dataset is obtained from the Kyoto Free Translation Task (KFTT) word alignment data\cite{neubig2011pointwise}, in which the sentences with less than 1 or more than 40 words are removed. The Japanese sentences are tokenized by KyTea tokenizer\cite{neubig2011pointwise}.

Regarding the datasets for our fine-tuning approach, we only use the first 80,000 sentence pairs of the whole training corpus for each language pair. Basically, the third-party supervision for these sentence pairs are extracted from the word alignments of the whole training corpus induced by the third-party aligner. We also test training the third-party aligner just on the 80,000 sentence pairs to provide the third-party supervision, the results are presented in section \ref{sec:retrain3rd}. Besides, we also vary the data size for the fine-tuning as shown in the experimental section \ref{sec:data_size}.

Table \ref{tab:datasize} presents the statistics of these datasets. Since De-En, En-Fr, and En-Ro have no manually aligned development sets, we take the last 1,000 sentences of the training data as the development sets\cite{ding2019saliency}, in which the aligner is self-tuned on the alignments predicted by itself in the last iteration. Other development sets and all test sets are manually aligned. All training sets do not contain manually labeled word alignments.

\begin{table}
\small
\centering
\begin{tabular}{lllll}
\hline
 & TRAIN & FINE-TUNE & DEV & TEST \\
 
\hline
Zh-En & 1,252,977 & 80,000 & 450 & 450 \\
De-En & 1,918,317 & 80,000 & 1,000 & 508  \\
En-Fr & 1,129,104 & 80,000 & 1,000 & 447 \\
Ro-En & 447,856 & 80,000 & 1,000 & 248 \\
Ja-En & 329,882 & 80,000 & 653 & 582 \\

\hline
\end{tabular}
\caption{\label{tab:datasize}
Number of sentence pairs in the benchmark datasets.
}
\end{table}

\subsection{Settings}

\paragraph{Pre-trained Cross-lingual Language Models.} For fine-tuning, we investigate two types of pre-trained CLMs, namely mBERT and XLM\cite{conneau2019cross}. mBERT is pre-trained over Wikipedia texts of 104 languages with the same settings to \citet{dou2021word}. For XLM, we have tried its two released models: 1) XLM-15(MLM+TLM) which is pre-trained with MLM and TLM objectives and supports 15 languages. 2) XLM-100(MLM), which is trained with MLM and supports 100 languages. Specifically, for Zh-En, De-En, and En-Fr, which are 
among the 15 languages, we use XLM-15(MLM+TLM) same to \citet{dou2021word}. For Ro-En and Ja-En which are not covered by XLM-15(MLM+TLM), we choose XLM-100 (MLM) instead with a modification that XLM-100 (MLM) is further trained on the parallel training corpora of Ro-en and Ja-En with the TLM objectives to be consistent with XLM-15(MLM+TLM). In the following, unless with clear specification, XLM stands for XLM-15 or XLM-100 in appropriate circumstances.

The contextualized word embeddings are extracted from the hidden states of the $i$th layer of the pre-trained CLMs, where $i$ is an empirically-chosen hyper-parameter based on the development set performances. For XLM-15, we use its 5th layer to extract the contextual embeddings \cite{hewitt2019structural,tenney2019bert}, while for XLM-100, we use its 9th layer. For mBERT, we use its 8th layer. We directly use the subwords in the pre-trained CLMs, i.e., BPE subwords in XLM and word piece subwords in mBERT . 

\paragraph{Training Setup and Hyper-parameters.} We fine-tune XLM and mBERT models for 10 epochs over the parallel fine-tuning corpus for each language pair, with a batch size of 8. We use AdamW\cite{loshchilov2017decoupled} with learning rate of 1e-5. The dropout rate is set to 0.1. The training process typically takes 2 to 3 hours. The hyper-parameters are tuned based on the development set performances. Regarding the threshold $c$ in the word alignment prediction, it is set to 1e-6 for Ro-En and 0.1 for the others. Regarding the hyper-parameters in integrating the various third-party supervisions, $f$ is set to 0.45 and $\lambda$ is set to 0.5 for all language pairs.

\subsection{Third-Party Word Aligners}

We explore various third-party word aligners ranging from statistical approaches to neural approaches to supervise the fine-tuning process. The aligners include:

\begin{itemize}

\item \textbf{FastAlign} \cite{dyer2013simple}\footnote{\url{https://github.com/clab/fast_align}}: a popular statistical word aligner which is an effective re-parameterization of IBM model 2. 

\item \textbf{GIZA++}\cite{och2003systematic}\footnote{\url{https://github.com/moses-smt/mgiza}}: another popular statistical word aligner implementing the IBM models. We use traditional settings of 5 iterations each for model 1, HMM model, model 3 and model 4. 

\item \textbf{Eflomal}\cite{ostling2016efficient}\footnote{\url{https://github.com/robertostling/eflomal}}: an efficient statistical word aligner using a Bayesian model with Markov Chain Monte Carlo inference.

\item \textbf{SimAlign}\cite{sabet2020simalign}\footnote{\url{https://github.com/cisnlp/simalign}}: a word aligner that directly uses static and contextualized embeddings of BERT to extract word alignments. We use its Argmax model with default settings. 

\item \textbf{AwesomeAlign}\cite{dou2021word}\footnote{\url{https://github.com/neulab/awesome-align}}: a neural word aligner that fine-tunes CLMs by self-training to produce contextualized embeddings suitable for word alignment.

\item \textbf{MaskAlign}\cite{chen2021mask}\footnote{\url{https://github.com/THUNLP-MT/Mask-Align}}: a neural word aligner based on self-supervision which parallel masks each target token and predicts it conditioned on both sides remaining tokens to better model the alignment.

\end{itemize}

For some language pairs that are not reported in the papers of the above third-party aligners, we run their released tools on the benchmark datasets to get the corresponding results. Specifically, for Zh-En, we run FastAlign, Eflomal, and SimAlign. For Ja-En, we run FastAlign, GIZA++, Eflomal, SimAlign, and MaskAlign. Because the evaluation in AwesomeAlign for Zh-En ignores manually labeled possible alignments, which is inconsistent to other works, we run AwesomeAlign for Zh-En to re-evaluate with considering the manually labeled possible alignments. 

\begin{table*}[h!]
\small
\setlength\tabcolsep{9pt}
\centering
\begin{tabular}{lcccccc}
\bottomrule[1.2pt]
& Zh-En & De-En & En-Fr & Ro-En & Ja-En & AVG \\
\hline \hline
\multicolumn{7}{c}{Baseline} \\
\hline
FastAlign \cite{dyer2013simple} &27.3 & 27.0 & 10.5 & 32.1 & 51.1 & 29.6\\
GIZA++ \cite{och2003systematic} & 18.5 & 20.6 & 5.9 & 26.4 & 48.0  & 23.9\\
Eflomal \cite{ostling2016efficient} & 23.4 & 22.6 &  8.2 & 25.1 & 47.5 & 25.4\\
SimAlign \cite{sabet2020simalign} & 19.6 & 19.0 & 6.0 & 30.5 & 48.6 & 26.4\\ 
AwesomeAlign \cite{dou2021word} &13.3 & 15.6 & 4.4 & 23.0 & 38.4 & 18.9 \\
MaskAlign \cite{chen2021mask} & 13.8 & 14.4 & 4.4 & 19.5 & 40.8 & 18.6\\
\hline \hline
\multicolumn{7}{c}{Fine-tuning XLM} \\
\hline
w/o Fine-tuning & 18.0 & 16.2 & 4.9 & 27.1 & 42.8 & 21.8\\
FastAlign$_{\rm adapted}$ & 23.0 & 27.0 & 11.2 & 32.2 & 49.3 & 28.5 \\
w/ FastAlign$_{\rm adapted}$ Supervision & 21.4 & 24.2 & 9.8 & 27.4 & 46.6 & 25.9\\
GIZA++$_{\rm adapted}$ & 18.8 & 19.3 & 6.3 & 29.0 & 43.9 & 23.5\\
w/ GIZA++$_{\rm adapted}$ Supervision & 13.3 & 15.2 & 5.1 & 23.8 & 39.2 & 19.3 \\
Eflomal$_{\rm adapted}$ & 27.0 & 26.0 &13.1 & 27.8 & 47.6 & 28.3  \\
w/ Eflomal$_{\rm adapted}$ Supervision & 14.0 & 18.4 & 6.1 & 23.6 & 43.7 & 21.2\\
SimAlign$_{\rm adapted}$ & 21.3 & 17.3 & 5.1 & 33.3 & 48.2 & 25.0\\
w/ SimAlign$_{\rm adapted}$ Supervision  & 14.7 & 14.8 & 4.5 & 26.5 & 44.0 & 20.9\\
AwesomeAlign$_{\rm adapted}$ & 13.7 & 17.2 & 4.7 & 24.2 & 40.4 & 20.0\\
w/ AwesomeAlign$_{\rm adapted}$ Supervision & 13.6 & 17.4 & 4.6 & 24.4 & 40.2 & 20.0\\
MaskAlign$_{\rm adapted}$ & 15.7 & 15.3 & 4.6 & 19.2 & 41.6 & 19.3\\
w/ MaskAlign$_{\rm adapted}$ Supervision & 12.1 & \textbf{13.9} & 4.3 & 18.8 & 34.3 & 16.7\\
\hline
w/ Integrated Supervision  & \textbf{11.3} & \textbf{13.9} & \textbf{4.0} & \textbf{18.6} & \textbf{33.4} & \textbf{16.2}\\
\bottomrule[1.2pt]
\end{tabular}
\caption{\label{tbl:main-results}
AER results of the baseline systems and the systems of fine-tuning XLM with the third-party supervisions. The lower AER, the better. AVG denotes the average AER over the five language pairs.
}
\end{table*}

\subsection{Main Results}

The alignment error rate (AER) \cite{och2003systematic} is used to evaluate the performances. Main results are summarized in Table \ref{tbl:main-results}. Compared to all third-party word aligners, which are also set as the baselines, our proposed approach achieves the state-of-the-art performances across the five language pairs, with an average AER of more than two points lower than the best third-party word aligner. 

Table \ref{tbl:main-results} presents the results of fine-tuning XLM. The results of fine-tuning mBERT is reported in Table \ref{tbl:main-results-mbert}. Both fine-tuning approaches perform better than the third-party word aligners. Since fine-tuning CLMs is conducted in the subword level, we need to adapt the third-party aligners for subwords. Given the parallel corpus of each language pair, we directly use the dictionary of the CLM to get the subwords of the corpus, then run each third-party aligner on such corpus which is subword segmented. Such adapted results are reported in both tables with the subscript ``adapted'' to each third-party aligner\footnote{We have tried other complicated adaptation approaches such as decomposing word alignments into subword alignments, adding pooling layers that deal with word level alignments, but they are not as effective as the above simple adaptation approach.}. For neural aligners such as MaskAlign which already uses subwords, the adaptation is still needed since the subwords of the pre-trained CLM are different. 

\begin{table*}[h!]
\small
\setlength\tabcolsep{9pt}
\centering
\begin{tabular}{lcccccc}
\bottomrule[1.2pt]
& Zh-En & De-En & En-Fr & Ro-En & Ja-En & AVG \\
\hline
\multicolumn{7}{c}{Fine-tuning mBERT} \\
\hline
w/o Fine-tuning & 17.9 & 17.4 & 5.6 & 27.3 & 45.2 & 22.7\\
FastAlign$_{\rm adapted}$ & 22.9 & 27.2 & 11.7 & 31.9 & 49.0 & 28.5\\
w/ FastAlign$_{\rm adapted}$ Supervision & 21.1 & 26.3 & 10.1 & 25.9 & 47.0 & 26.1\\
Eflomal$_{\rm adapted}$ & 27.2 & 25.9 & 13.0 & 26.8 & 48.5 & 28.3 \\
w/ Eflomal$_{\rm adapted}$ Supervision & 16.3 & 21.0 &7.2 & 22.3 & 44.4 & 22.2\\
GIZA++$_{\rm adapted}$ & 18.3 & 19.9 & 6.3 & 27.6 & 42.6 & 22.9\\
w/ GIZA++$_{\rm adapted}$ Supervision  & 13.5 & 17.5 & 5.2 & 23.2 & 37.7 & 19.4 \\
SimAlign$_{\rm adapted}$ & 19.6 & 19.0 & 5.9 & 30.5 & 48.6 & 24.7\\
w/ SimAlign$_{\rm adapted}$ Supervision  & 16.6 & 16.2 & 5.4 & 24.0 & 43.9 & 21.2\\
AwesomeAlign$_{\rm adapted}$ & 13.3 & 15.2 & 4.3 & 23.3 & 38.5 & 18.9\\
w/ AwesomeAlign$_{\rm adapted}$ Supervision & 13.4 & 15.0 & 4.5 & 23.0 & 38.2 & 18.8\\
MaskAlign$_{\rm adapted}$ & 15.7 & 15.9 & 4.3 & 20.3 & 41.6 & 19.6 \\
w/ MaskAlign$_{\rm adapted}$ Supervision  & 11.5 & 15.2 &3.9 & 19.5 & 34.6 & 16.9\\
\hline
w/ Integrated Supervision  & \textbf{11.0} & \textbf{14.8} & \textbf{3.8} & \textbf{19.3} & \textbf{33.2} & \textbf{16.4}\\
\bottomrule[1.2pt]
\end{tabular}
\caption{\label{tbl:main-results-mbert}
AER results of fine-tuning mBERT with the third-party supervisions. 
}
\end{table*}

Regarding the plain contextualized embeddings in XLM and mBERT, they can be directly aligned between source and target languages by mining the closest neighbors in the universal embedding space, as shown in the ``w/o Fine-tuning'' rows in Table \ref{tbl:main-results} and \ref{tbl:main-results-mbert} \cite{dou2021word}. When we further fine-tune these embeddings supervised by the subword alignments produced by each adapted individual third-party aligner, we obtain significant improvement over each individual third-party aligner. When compare fine-tuning to without fine-tuning (``w/o Fine-tuning'' rows), we found that fine-tuning generally performs better than without fine-tuning, except for fine-tuning with the supervision of FastAlign$_{\rm adapted}$. Since FastAlign$_{\rm adapted}$ performs remarkably worse than without fine-tuning, it is hard for FastAlign$_{\rm adapted}$ to provide effective supervision for the fine-tuning. Since AwsomeAlign$_{\rm adapted}$ already fine-tunes the CLMs by self-training, continuing to fine-tune CLMs with the supervision of AwsomeAlign$_{\rm adapted}$ does not gain improvements. At last, when we integrate all supervisions from various third-party aligners, we achieve state-of-the-art AER. Details of integrating all supervisions are presented in section \ref{sec:integration}.

\begin{figure}[tb]
    \centering
    \includegraphics[width=8cm]{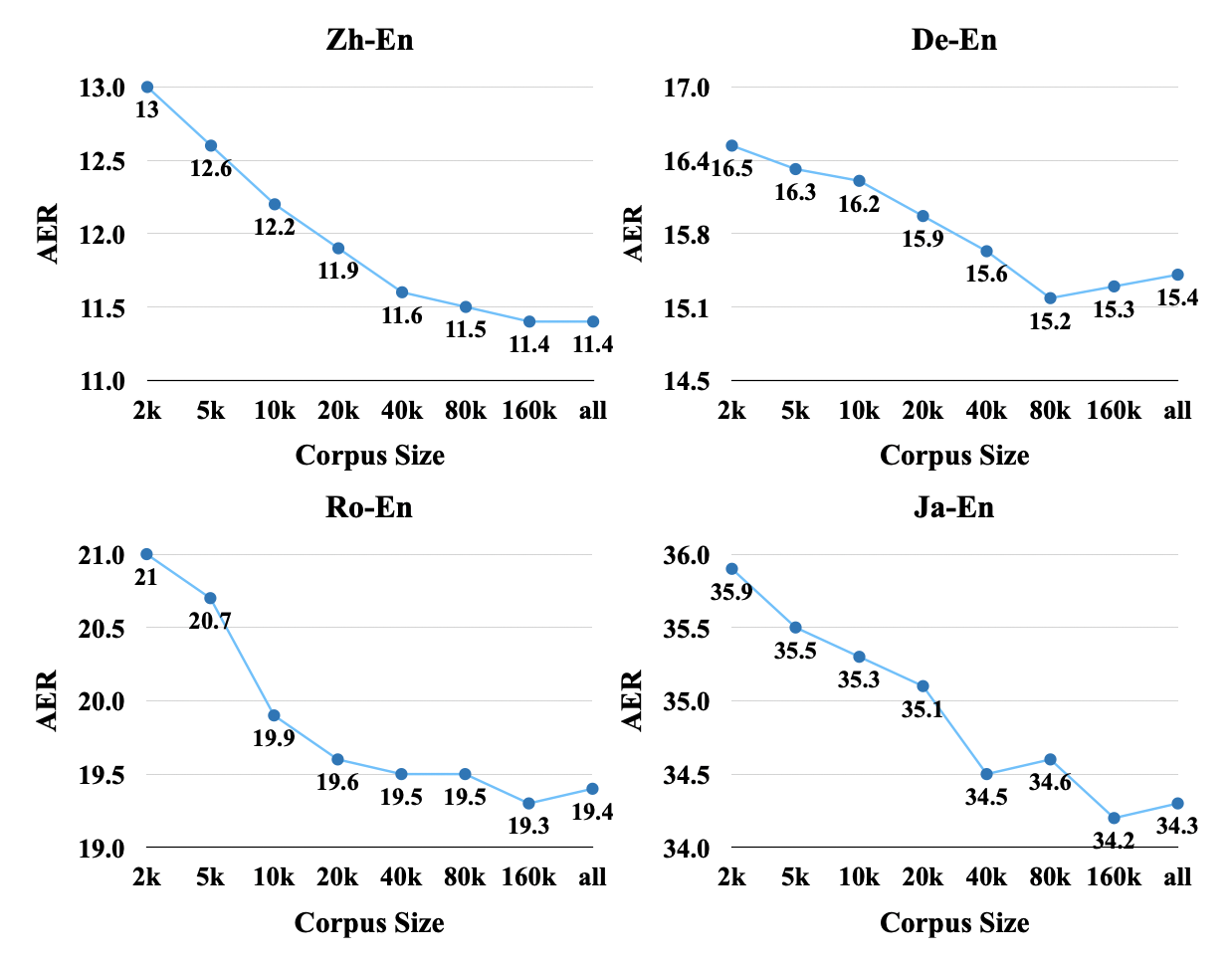}
    \caption{The effect of the different sizes of parallel corpora for the fine-tuning.}
    \label{fig:datasize-for-finetune}
\end{figure}

\subsection{The Effect of The Fine-tuning Corpus Size} \label{sec:data_size}

Figure \ref{fig:datasize-for-finetune} presents the performance variance when the size of parallel corpus for the fine-tuning varies. As the fine-tuning corpus becomes larger, AER becomes lower across all five language pairs. The full corpus is identical to that used in training the third-party aligners. The curve for En-Fr is presented in the appendix due to space limit. Usually 80k sentence pairs can provide good supervisions for the fine-tuning, with limited margin to the performance of using the full corpus. Note that the performance of using 2k sentence pairs for fine-tuning is less than two points worse than that of using the full corpus, even just 0.4 points worse in En-Fr. 

\subsection{Self-Correction Effect} \label{sec:self_correction}

Although the supervision from the third-party aligner is not perfect, we observe a self-correction effect that as the fine-tuning proceeds, more accurate word alignments other than the third-party alignments are identified as they become closer in the embedding space, and some wrong word alignments of the third-party aligner get departed farther in the space, which we deem that they do not influence the fine-tuning process.

\begin{figure}[tb]
    \centering
    \includegraphics[width=8cm]{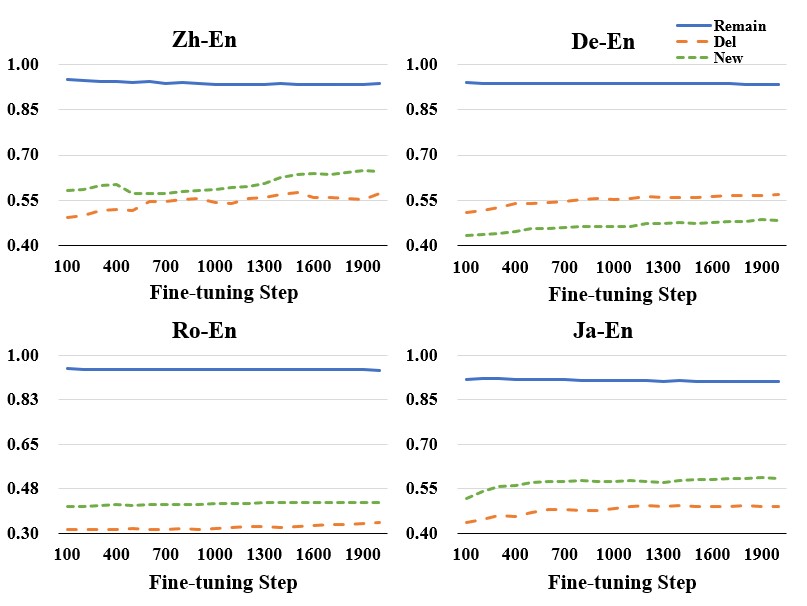}
    \caption{Self-correction effect in the fine-tuning process.}
    \label{fig:self-correction-process}
\end{figure}

Figure \ref{fig:self-correction-process} presents the self-correction effect. In this subsection, we include the test set into the fine-tuning set for the new fine-tuning to check the predicted alignments against gold alignments. MaskAlign and XLM are used in this study. At first, we extract MaskAlign results of the test set as part of the supervision for the fine-tuning. As the fine-tuning steps forward, on the test set, we compute the precision of newly predicted alignments not included in the third-party alignments, denoted as ``New'', and the rate of the deleted alignments (certain third-party alignments not included in the predicted alignments) which are truly wrong alignments amongst all deleted alignments, denoted as ``Del''. Besides, we compute the precision of remaining alignments in the third-party alignments, denoted as ``Remain''. Figure \ref{fig:self-correction-process} shows that ``New'' and ``Del'' increase as the fine-tuning proceeds, supporting the AER decrease in the experiment. ``Remain'' almost keeps horizontal, indicating the stability of the fine-tuning process. The effect of En-Fr is shown in the appendix.

\subsection{Results of Integrating Various Third-Party Supervisions} \label{sec:integration}

Table~\ref{tbl:ensemble} presents the comparison between the performances of different integration methods. Fine-tuning mBERT is applied in this study for its computation efficiency. First, we intersect the word alignments from all third-party aligners as supervisions. Since aligners perform differently, AER is impacted by the worst aligner which results in small number of word alignments in the intersection. In contrast, when we get the union of all third-party alignments, its performance is much better, but it still contain noises hampering AER results. When we use filtering and weighting methods to deal with the noises, the integration gets the best performances, and surpasses all third-party aligners.

\begin{table}
\setlength\tabcolsep{2pt}
\small
\centering
\begin{tabular}{lcccccc}
\hline
& Zh-En & De-En & En-Fr & Ro-En & Ja-En & AVG\\
\hline
 Intersection & 13.0 & 17.0 & 4.8 & 23.4 & 39.4 & 19.5\\
 Union & 11.8 & 15.2 & 4.1 & 20.4 & 35.1& 17.3\\
 Union-Filtering & 11.0 & 14.8 & 3.8 & 19.3 & 33.2 & 16.4\\
 Union-Weighting & 11.2 & 14.7 & 3.8 & 19.2 & 33.7 & 16.5\\
\hline
\end{tabular}
\caption{\label{tbl:ensemble}
 AER of different integration methods.
}
\end{table}

Ablation studies are shown in Table \ref{tbl:ablation}. Removing one aligner from the integration causes different performance variances. It shows that removing MaskAlign impact the integration performance most, since it is best aligner in most language pairs. 

\begin{table}
\setlength\tabcolsep{1.8pt}
\small
\centering
\begin{tabular}{lcccccc}
\hline
& Zh-En & De-En & En-Fr & Ro-En & Ja-En & AVG\\
\hline
 w/o FastAlign & 11.2 & 14.9 & 3.9 & 19.1 & 33.7 & 16.5\\
 w/o Eflomal & 11.0 & 15.1 & 4.0 & 19.6 & 34.0 & 16.7\\
 w/o GIZA++ & 11.2 & 15.1 & 3.8 & 19.4 & 35.2& 16.7\\
 w/o SimAlign & 11.3 & 15.0 & 3.9 & 19.3 & 33.8 & 16.6\\
 w/o MaskAlign & 12.3 & 15.1 & 4.2 & 22.5 & 36.4 & 18.1\\
 w/o AwesomeAlign & 11.5 & 15.4 & 4.0 & 19.4 & 34.4 & 16.9\\
 All & 11.0 & 14.8 & 3.8 & 19.3 & 33.2 &16.4\\
\hline
\end{tabular}
\caption{\label{tbl:ablation}
Ablation studies of the integration method using Union-Filtering.
}
\end{table}

\begin{table}
\setlength\tabcolsep{2pt}
\small
\centering
\begin{tabular}{lcccccc}
\hline
 & Zh-En & De-En & En-Fr & Ro-En & Ja-En & AVG\\
\hline
\multicolumn{7}{c}{XLM} \\
\hline
MaskAlign$_{\rm adapted}$ & 23.8 & 27.8 & 7.5 & 23.1 & 66.1 & 29.7 \\
Fine-tuning &  13.6 & 14.5 & 4.2 & 20.9 & 36.9 &  18.0 \\
\hline
\multicolumn{7}{c}{mBERT} \\
\hline
MaskAlign$_{\rm adapted}$ & 19.8 & 25.5 & 7.6 & 20.3 & 61.0 & 26.8 \\
Fine-tuning & 11.6 & 14.7 & 4.8 & 19.9 & 35.3 & 17.3  \\
\hline
\end{tabular}
\caption{\label{tbl:8w-maskalign}
AER of fine-tuning XLM and mBERT with the third-party supervision, which is generated by MaskAlign$_{\rm adapted}$ trained on the small parallel corpus same to that used in the fine-tuning.
}
\end{table}

\subsection{Training Third-Party Aligners on The Same Parallel Corpus for The Fine-tuning} \label{sec:retrain3rd}

Although the fine-tuning approach only needs a small fraction of the whole parallel corpus for each language pair, e.g. 80k sentence pairs for the fine-tuning, its supervision is extracted from the alignments of the third-party aligner which is trained on the whole parallel corpus. In this subsection, we check if only using the small corpus, which is used in the fine-tuning, for training the third-party aligner can seriously impact the word alignment performance. Table \ref{tbl:8w-maskalign} shows the result. Training MaskAlign on small corpus seriously drags down AER performances when compared to training on full corpus, with averagely over 7 points worse than ``MaskAlign$_{\rm adapted}$'' in Table \ref{tbl:main-results} and \ref{tbl:main-results-mbert}. Surprisingly, fine-tuning with such worse supervision can still achieve remarkably better performances, even surpassing or performing comparable to the strongest baseline system MaskAlign in Table \ref{tbl:main-results}. The reason for this phenomenon is that MaskAlign$_{\rm adapted}$ generates fewer but more accurate alignments, which is effective enough for the supervision. We also use 40k sentence pairs for this study. Please refer to Appendix C for the study. 

\section{Conclusion}
We propose an approach of using a third-party aligner for neural word alignments. Different to previous work based on careful design of self-training objectives, we simply use the word alignments generated by the third-party aligners to supervise the training. Although the third-party word alignments are imperfect as the supervision, we observe that the training process can do self-correction over the third-party word alignments by detecting more accurate word alignments and deleting wrong word alignments based on the geometry similarity in the contextualized embedding space, leading to better performances than the third-party aligners. The integration of various third-party supervisions improves the performance further, achieving state-of-the-art word alignment performance on benchmarks of multiple language pairs.

\section*{Limitations}

The proposed third-party supervised fine-tuning approach is not applicable to using the best word alignments, which are generated by the integrated supervision in this paper, as the new supervision signal to continue the fine-tuning. Such continual fine-tuning does not obtain significant improvement, which indicates the ineffectiveness of continual fine-tuning with the supervision of self predicted alignments.

\section*{Ethics Statement}

The data used in our experiments are either freely downloadable from web or obtained via the LDC license. The codes of the third-party aligners and the pre-trained CLMs are freely downloadable from web. 

\section*{Acknowledgments}
The authors would like to thank the anonymous reviewers for the helpful comments. This work was supported by National Natural Science Foundation of China (Grant No. 62276179, 62036004), and was also partially supported by the joint research project of Alibaba and Soochow University.

% Entries for the entire Anthology, followed by custom entries
\bibliography{anthology,custom}

\begin{thebibliography}{46}
\expandafter\ifx\csname natexlab\endcsname\relax\def\natexlab#1{#1}\fi

\bibitem[{Agi{\'c} et~al.(2016)Agi{\'c}, Johannsen, Plank, Alonso, Schluter,
  and S{\o}gaard}]{agic2016multilingual}
{\v{Z}}eljko Agi{\'c}, Anders Johannsen, Barbara Plank,
  H{\'e}ctor~Mart{\'\i}nez Alonso, Natalie Schluter, and Anders S{\o}gaard.
  2016.
\newblock Multilingual projection for parsing truly low-resource languages.
\newblock \emph{Transactions of the Association for Computational Linguistics},
  4:301--312.

\bibitem[{Ammar et~al.(2016)Ammar, Mulcaire, Tsvetkov, Lample, Dyer, and
  Smith}]{Ammar2016}
Waleed Ammar, George Mulcaire, Yulia Tsvetkov, Guillaume Lample, Chris Dyer,
  and Noah~A. Smith. 2016.
\newblock \href {https://arxiv.org/pdf/1602.01925.pdf} {Massively multilingual
  word embeddings}.
\newblock \emph{arXiv 1602.01925}.

\bibitem[{Arthur et~al.(2016)Arthur, Neubig, and Nakamura}]{arthur2016}
Philip Arthur, Graham Neubig, and Satoshi Nakamura. 2016.
\newblock \href {https://doi.org/10.18653/v1/D16-1162} {Incorporating discrete
  translation lexicons into neural machine translation}.
\newblock In \emph{Proceedings of the 2016 Conference on Empirical Methods in
  Natural Language Processing}, pages 1557--1567, Austin, Texas. Association
  for Computational Linguistics.

\bibitem[{Bau et~al.(2018)Bau, Belinkov, Sajjad, Durrani, Dalvi, and
  Glass}]{bau2018identifying}
Anthony Bau, Yonatan Belinkov, Hassan Sajjad, Nadir Durrani, Fahim Dalvi, and
  James Glass. 2018.
\newblock Identifying and controlling important neurons in neural machine
  translation.
\newblock In \emph{Proceedings of the International Conference on Learning
  Representations}.

\bibitem[{Brown et~al.(1993)Brown, Della~Pietra, Della~Pietra, and
  Mercer}]{brown1993}
Peter~F. Brown, Stephen~A. Della~Pietra, Vincent~J. Della~Pietra, and Robert~L.
  Mercer. 1993.
\newblock \href {https://aclanthology.org/J93-2003} {The mathematics of
  statistical machine translation: Parameter estimation}.
\newblock \emph{Computational Linguistics}, 19(2):263--311.

\bibitem[{Cao et~al.(2019)Cao, Kitaev, and Klein}]{cao2019multilingual}
Steven Cao, Nikita Kitaev, and Dan Klein. 2019.
\newblock Multilingual alignment of contextual word representations.
\newblock In \emph{Proceedings of the International Conference on Learning
  Representations}.

\bibitem[{Chen et~al.(2021)Chen, Sun, and Liu}]{chen2021mask}
Chi Chen, Maosong Sun, and Yang Liu. 2021.
\newblock \href {https://doi.org/10.18653/v1/2021.acl-long.369} {Mask-align:
  Self-supervised neural word alignment}.
\newblock In \emph{Proceedings of the 59th Annual Meeting of the Association
  for Computational Linguistics and the 11th International Joint Conference on
  Natural Language Processing (Volume 1: Long Papers)}, pages 4781--4791,
  Online. Association for Computational Linguistics.

\bibitem[{Chen et~al.(2020)Chen, Liu, Chen, Jiang, and Liu}]{chen2020accurate}
Yun Chen, Yang Liu, Guanhua Chen, Xin Jiang, and Qun Liu. 2020.
\newblock \href {https://doi.org/10.18653/v1/2020.emnlp-main.42} {Accurate word
  alignment induction from neural machine translation}.
\newblock In \emph{Proceedings of the 2020 Conference on Empirical Methods in
  Natural Language Processing (EMNLP)}, pages 566--576, Online. Association for
  Computational Linguistics.

\bibitem[{Conneau et~al.(2020)Conneau, Khandelwal, Goyal, Chaudhary, Wenzek,
  Guzm{\'a}n, Grave, Ott, Zettlemoyer, and Stoyanov}]{conneau2020unsupervised}
Alexis Conneau, Kartikay Khandelwal, Naman Goyal, Vishrav Chaudhary, Guillaume
  Wenzek, Francisco Guzm{\'a}n, Edouard Grave, Myle Ott, Luke Zettlemoyer, and
  Veselin Stoyanov. 2020.
\newblock \href {https://doi.org/10.18653/v1/2020.acl-main.747} {Unsupervised
  cross-lingual representation learning at scale}.
\newblock In \emph{Proceedings of the 58th Annual Meeting of the Association
  for Computational Linguistics}, pages 8440--8451, Online. Association for
  Computational Linguistics.

\bibitem[{Conneau and Lample(2019)}]{conneau2019cross}
Alexis Conneau and Guillaume Lample. 2019.
\newblock Cross-lingual language model pretraining.
\newblock \emph{Advances in neural information processing systems}, 32.

\bibitem[{Devlin et~al.(2019)Devlin, Chang, Lee, and
  Toutanova}]{devlin2019bert}
Jacob Devlin, Ming-Wei Chang, Kenton Lee, and Kristina Toutanova. 2019.
\newblock \href {https://doi.org/10.18653/v1/N19-1423} {{BERT}: Pre-training of
  deep bidirectional transformers for language understanding}.
\newblock In \emph{Proceedings of the 2019 Conference of the North {A}merican
  Chapter of the Association for Computational Linguistics: Human Language
  Technologies, Volume 1 (Long and Short Papers)}, pages 4171--4186,
  Minneapolis, Minnesota. Association for Computational Linguistics.

\bibitem[{Ding et~al.(2019)Ding, Xu, and Koehn}]{ding2019saliency}
Shuoyang Ding, Hainan Xu, and Philipp Koehn. 2019.
\newblock \href {https://doi.org/10.18653/v1/W19-5201} {Saliency-driven word
  alignment interpretation for neural machine translation}.
\newblock In \emph{Proceedings of the Fourth Conference on Machine Translation
  (Volume 1: Research Papers)}, pages 1--12, Florence, Italy. Association for
  Computational Linguistics.

\bibitem[{Dou and Neubig(2021)}]{dou2021word}
Zi-Yi Dou and Graham Neubig. 2021.
\newblock \href {https://doi.org/10.18653/v1/2021.eacl-main.181} {Word
  alignment by fine-tuning embeddings on parallel corpora}.
\newblock In \emph{Proceedings of the 16th Conference of the European Chapter
  of the Association for Computational Linguistics: Main Volume}, pages
  2112--2128, Online. Association for Computational Linguistics.

\bibitem[{Dyer et~al.(2013)Dyer, Chahuneau, and Smith}]{dyer2013simple}
Chris Dyer, Victor Chahuneau, and Noah~A Smith. 2013.
\newblock A simple, fast, and effective reparameterization of ibm model 2.
\newblock In \emph{Proceedings of the 2013 Conference of the North American
  Chapter of the Association for Computational Linguistics: Human Language
  Technologies}, pages 644--648.

\bibitem[{Garg et~al.(2019)Garg, Peitz, Nallasamy, and
  Paulik}]{garg2019jointly}
Sarthak Garg, Stephan Peitz, Udhyakumar Nallasamy, and Matthias Paulik. 2019.
\newblock \href {https://doi.org/10.18653/v1/D19-1453} {Jointly learning to
  align and translate with transformer models}.
\newblock In \emph{Proceedings of the 2019 Conference on Empirical Methods in
  Natural Language Processing and the 9th International Joint Conference on
  Natural Language Processing (EMNLP-IJCNLP)}, pages 4453--4462, Hong Kong,
  China. Association for Computational Linguistics.

\bibitem[{Hasler et~al.(2018)Hasler, de~Gispert, Iglesias, and
  Byrne}]{hasler2018}
Eva Hasler, Adri{\`a} de~Gispert, Gonzalo Iglesias, and Bill Byrne. 2018.
\newblock \href {https://doi.org/10.18653/v1/N18-2081} {Neural machine
  translation decoding with terminology constraints}.
\newblock In \emph{Proceedings of the 2018 Conference of the North {A}merican
  Chapter of the Association for Computational Linguistics: Human Language
  Technologies, Volume 2 (Short Papers)}, pages 506--512, New Orleans,
  Louisiana. Association for Computational Linguistics.

\bibitem[{Herzig and Berant(2018)}]{herzig2018}
Jonathan Herzig and Jonathan Berant. 2018.
\newblock \href {https://doi.org/10.18653/v1/D18-1190} {Decoupling structure
  and lexicon for zero-shot semantic parsing}.
\newblock In \emph{Proceedings of the 2018 Conference on Empirical Methods in
  Natural Language Processing}, pages 1619--1629, Brussels, Belgium.
  Association for Computational Linguistics.

\bibitem[{Hewitt and Manning(2019)}]{hewitt2019structural}
John Hewitt and Christopher~D Manning. 2019.
\newblock A structural probe for finding syntax in word representations.
\newblock In \emph{Proceedings of the 2019 Conference of the North American
  Chapter of the Association for Computational Linguistics: Human Language
  Technologies, Volume 1 (Long and Short Papers)}, pages 4129--4138.

\bibitem[{Hu et~al.(2020)Hu, Ruder, Siddhant, Neubig, Firat, and
  Johnson}]{hu2020xtreme}
Junjie Hu, Sebastian Ruder, Aditya Siddhant, Graham Neubig, Orhan Firat, and
  Melvin Johnson. 2020.
\newblock Xtreme: A massively multilingual multi-task benchmark for evaluating
  cross-lingual generalisation.
\newblock In \emph{International Conference on Machine Learning}, pages
  4411--4421. PMLR.

\bibitem[{Li et~al.(2019)Li, Li, Liu, Meng, and Shi}]{li2019word}
Xintong Li, Guanlin Li, Lemao Liu, Max Meng, and Shuming Shi. 2019.
\newblock On the word alignment from neural machine translation.
\newblock In \emph{Proceedings of the 57th Annual Meeting of the Association
  for Computational Linguistics}, pages 1293--1303.

\bibitem[{Libovick{\`y} et~al.(2019)Libovick{\`y}, Rosa, and
  Fraser}]{libovicky2019language}
Jind{\v{r}}ich Libovick{\`y}, Rudolf Rosa, and Alexander Fraser. 2019.
\newblock How language-neutral is multilingual bert?
\newblock \emph{arXiv preprint arXiv:1911.03310}.

\bibitem[{Liu et~al.(2016)Liu, Utiyama, Finch, and Sumita}]{liu2016neural}
Lemao Liu, Masao Utiyama, Andrew Finch, and Eiichiro Sumita. 2016.
\newblock \href {https://aclanthology.org/C16-1291} {Neural machine translation
  with supervised attention}.
\newblock In \emph{Proceedings of {COLING} 2016, the 26th International
  Conference on Computational Linguistics: Technical Papers}, pages 3093--3102,
  Osaka, Japan. The COLING 2016 Organizing Committee.

\bibitem[{Liu et~al.(2005)Liu, Liu, and Lin}]{liu2005log}
Yang Liu, Qun Liu, and Shouxun Lin. 2005.
\newblock \href {https://doi.org/10.3115/1219840.1219897} {Log-linear models
  for word alignment}.
\newblock In \emph{Proceedings of the 43rd Annual Meeting of the Association
  for Computational Linguistics ({ACL}{'}05)}, pages 459--466, Ann Arbor,
  Michigan. Association for Computational Linguistics.

\bibitem[{Loshchilov and Hutter(2017)}]{loshchilov2017decoupled}
Ilya Loshchilov and Frank Hutter. 2017.
\newblock Decoupled weight decay regularization.
\newblock \emph{arXiv preprint arXiv:1711.05101}.

\bibitem[{Mayhew et~al.(2017)Mayhew, Tsai, and Roth}]{mayhew2017cheap}
Stephen Mayhew, Chen-Tse Tsai, and Dan Roth. 2017.
\newblock Cheap translation for cross-lingual named entity recognition.
\newblock In \emph{Proceedings of the 2017 conference on empirical methods in
  natural language processing}, pages 2536--2545.

\bibitem[{Neubig et~al.(2019)Neubig, Dou, Hu, Michel, Pruthi, and
  Wang}]{neubig2019compare}
Graham Neubig, Zi-Yi Dou, Junjie Hu, Paul Michel, Danish Pruthi, and Xinyi
  Wang. 2019.
\newblock \href {https://doi.org/10.18653/v1/N19-4007} {compare-mt: A tool for
  holistic comparison of language generation systems}.
\newblock In \emph{Proceedings of the 2019 Conference of the North {A}merican
  Chapter of the Association for Computational Linguistics (Demonstrations)},
  pages 35--41, Minneapolis, Minnesota. Association for Computational
  Linguistics.

\bibitem[{Neubig et~al.(2011)Neubig, Nakata, and Mori}]{neubig2011pointwise}
Graham Neubig, Yosuke Nakata, and Shinsuke Mori. 2011.
\newblock Pointwise prediction for robust, adaptable japanese morphological
  analysis.
\newblock In \emph{Proceedings of the 49th Annual Meeting of the Association
  for Computational Linguistics: Human Language Technologies}, pages 529--533.

\bibitem[{Nicolai and Yarowsky(2019)}]{nicolai2019learning}
Garrett Nicolai and David Yarowsky. 2019.
\newblock Learning morphosyntactic analyzers from the bible via iterative
  annotation projection across 26 languages.
\newblock In \emph{Proceedings of the 57th Annual Meeting of the Association
  for Computational Linguistics}, pages 1765--1774.

\bibitem[{Och and Ney(2003)}]{och2003systematic}
Franz~Josef Och and Hermann Ney. 2003.
\newblock A systematic comparison of various statistical alignment models.
\newblock \emph{Computational linguistics}, 29(1):19--51.

\bibitem[{{\"O}stling and Tiedemann(2016)}]{ostling2016efficient}
Robert {\"O}stling and J{\"o}rg Tiedemann. 2016.
\newblock Efficient word alignment with markov chain monte carlo.
\newblock \emph{The Prague Bulletin of Mathematical Linguistics}.

\bibitem[{Pad{\'o} and Lapata(2009)}]{pado2009}
Sebastian Pad{\'o} and Mirella Lapata. 2009.
\newblock Cross-lingual annotation projection for semantic roles.
\newblock \emph{Journal of Artificial Intelligence Research}, 36:307--340.

\bibitem[{Pal et~al.(2017)Pal, Naskar, Vela, Liu, and van Genabith}]{pal2017}
Santanu Pal, Sudip~Kumar Naskar, Mihaela Vela, Qun Liu, and Josef van Genabith.
  2017.
\newblock \href {https://aclanthology.org/E17-2056} {Neural automatic
  post-editing using prior alignment and reranking}.
\newblock In \emph{Proceedings of the 15th Conference of the {E}uropean Chapter
  of the Association for Computational Linguistics: Volume 2, Short Papers},
  pages 349--355, Valencia, Spain. Association for Computational Linguistics.

\bibitem[{Peters et~al.(2019)Peters, Niculae, and Martins}]{peters2019sparse}
Ben Peters, Vlad Niculae, and Andr{\'e} F.~T. Martins. 2019.
\newblock \href {https://doi.org/10.18653/v1/P19-1146} {Sparse
  sequence-to-sequence models}.
\newblock In \emph{Proceedings of the 57th Annual Meeting of the Association
  for Computational Linguistics}, pages 1504--1519, Florence, Italy.
  Association for Computational Linguistics.

\bibitem[{Sabet et~al.(2020)Sabet, Dufter, Yvon, and
  Sch{\"u}tze}]{sabet2020simalign}
Masoud~Jalili Sabet, Philipp Dufter, Fran{\c{c}}ois Yvon, and Hinrich
  Sch{\"u}tze. 2020.
\newblock \href {https://doi.org/10.18653/v1/2020.findings-emnlp.147}
  {{S}im{A}lign: High quality word alignments without parallel training data
  using static and contextualized embeddings}.
\newblock In \emph{Findings of the Association for Computational Linguistics:
  EMNLP 2020}, pages 1627--1643, Online. Association for Computational
  Linguistics.

\bibitem[{Sennrich et~al.(2016)Sennrich, Haddow, and
  Birch}]{sennrich2016neural}
Rico Sennrich, Barry Haddow, and Alexandra Birch. 2016.
\newblock \href {https://doi.org/10.18653/v1/P16-1162} {Neural machine
  translation of rare words with subword units}.
\newblock In \emph{Proceedings of the 54th Annual Meeting of the Association
  for Computational Linguistics (Volume 1: Long Papers)}, pages 1715--1725,
  Berlin, Germany. Association for Computational Linguistics.

\bibitem[{Song et~al.(2020{\natexlab{a}})Song, Wang, Yu, Zhang, Huang, Luo,
  Duan, and Zhang}]{song2020alignment}
Kai Song, Kun Wang, Heng Yu, Yue Zhang, Zhongqiang Huang, Weihua Luo, Xiangyu
  Duan, and Min Zhang. 2020{\natexlab{a}}.
\newblock Alignment-enhanced transformer for constraining nmt with
  pre-specified translations.
\newblock In \emph{Proceedings of the AAAI Conference on Artificial
  Intelligence}, volume~34, pages 8886--8893.

\bibitem[{Song et~al.(2020{\natexlab{b}})Song, Zhou, Yu, Huang, Zhang, Luo,
  Duan, and Zhang}]{song2020towards}
Kai Song, Xiaoqing Zhou, Heng Yu, Zhongqiang Huang, Yue Zhang, Weihua Luo,
  Xiangyu Duan, and Min Zhang. 2020{\natexlab{b}}.
\newblock Towards better word alignment in transformer.
\newblock \emph{IEEE/ACM Transactions on Audio, Speech, and Language
  Processing}, 28:1801--1812.

\bibitem[{Stanovsky et~al.(2019)Stanovsky, Smith, and
  Zettlemoyer}]{stanovsky2019}
Gabriel Stanovsky, Noah~A. Smith, and Luke Zettlemoyer. 2019.
\newblock \href {https://doi.org/10.18653/v1/P19-1164} {Evaluating gender bias
  in machine translation}.
\newblock In \emph{Proceedings of the 57th Annual Meeting of the Association
  for Computational Linguistics}, pages 1679--1684, Florence, Italy.
  Association for Computational Linguistics.

\bibitem[{Tenney et~al.(2019)Tenney, Das, and Pavlick}]{tenney2019bert}
Ian Tenney, Dipanjan Das, and Ellie Pavlick. 2019.
\newblock Bert rediscovers the classical nlp pipeline.
\newblock \emph{arXiv preprint arXiv:1905.05950}.

\bibitem[{Tiedemann(2014)}]{tiedemann2014}
J{\"o}rg Tiedemann. 2014.
\newblock Rediscovering annotation projection for cross-lingual parser
  induction.
\newblock In \emph{Proceedings of COLING 2014, the 25th International
  Conference on Computational Linguistics: Technical Papers}, pages 1854--1864.

\bibitem[{Tu et~al.(2016)Tu, Lu, Liu, Liu, and Li}]{tu2016}
Zhaopeng Tu, Zhengdong Lu, Yang Liu, Xiaohua Liu, and Hang Li. 2016.
\newblock \href {https://doi.org/10.18653/v1/P16-1008} {Modeling coverage for
  neural machine translation}.
\newblock In \emph{Proceedings of the 54th Annual Meeting of the Association
  for Computational Linguistics (Volume 1: Long Papers)}, pages 76--85, Berlin,
  Germany. Association for Computational Linguistics.

\bibitem[{Wang et~al.(2020)Wang, Tu, Shi, and Liu}]{wang2020inference}
Shuo Wang, Zhaopeng Tu, Shuming Shi, and Yang Liu. 2020.
\newblock \href {https://doi.org/10.18653/v1/2020.acl-main.278} {On the
  inference calibration of neural machine translation}.
\newblock In \emph{Proceedings of the 58th Annual Meeting of the Association
  for Computational Linguistics}, pages 3070--3079, Online. Association for
  Computational Linguistics.

\bibitem[{Wu et~al.(2016)Wu, Schuster, Chen, Le, Norouzi, Macherey, Krikun,
  Cao, Gao, Macherey et~al.}]{wu2016google}
Yonghui Wu, Mike Schuster, Zhifeng Chen, Quoc~V Le, Mohammad Norouzi, Wolfgang
  Macherey, Maxim Krikun, Yuan Cao, Qin Gao, Klaus Macherey, et~al. 2016.
\newblock Google's neural machine translation system: Bridging the gap between
  human and machine translation.
\newblock \emph{arXiv preprint arXiv:1609.08144}.

\bibitem[{Yarowsky et~al.(2001)Yarowsky, Ngai, and Wicentowski}]{yarowsky2001}
David Yarowsky, Grace Ngai, and Richard Wicentowski. 2001.
\newblock \href {https://aclanthology.org/H01-1035} {Inducing multilingual text
  analysis tools via robust projection across aligned corpora}.
\newblock In \emph{Proceedings of the First International Conference on Human
  Language Technology Research}.

\bibitem[{Zenkel et~al.(2019)Zenkel, Wuebker, and DeNero}]{zenkel2019adding}
Thomas Zenkel, Joern Wuebker, and John DeNero. 2019.
\newblock Adding interpretable attention to neural translation models improves
  word alignment.
\newblock \emph{arXiv preprint arXiv:1901.11359}.

\bibitem[{Zenkel et~al.(2020)Zenkel, Wuebker, and DeNero}]{zenkel2020end}
Thomas Zenkel, Joern Wuebker, and John DeNero. 2020.
\newblock \href {https://doi.org/10.18653/v1/2020.acl-main.146} {End-to-end
  neural word alignment outperforms {GIZA}++}.
\newblock In \emph{Proceedings of the 58th Annual Meeting of the Association
  for Computational Linguistics}, pages 1605--1617, Online. Association for
  Computational Linguistics.

\end{thebibliography}
\bibliographystyle{acl_natbib}

\appendix

\section{The Corpus Size Effect and The Self-Correction Effect on En-Fr}

\begin{figure}[h!]
    \centering
    \includegraphics[width=7.5cm]{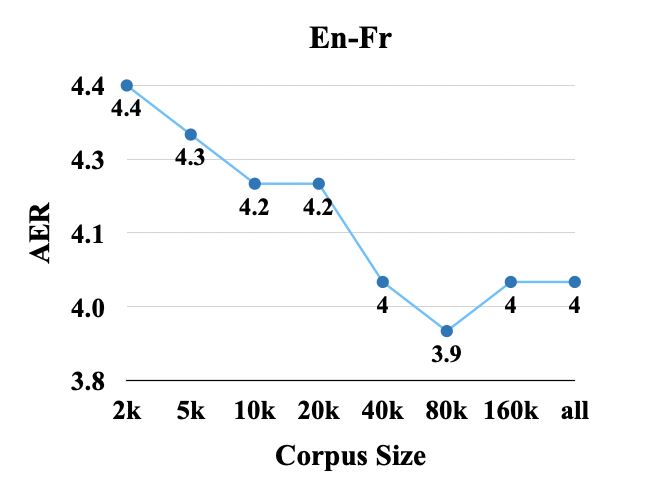}
    \caption{The effect of the different sizes of the parallel corpus for En-Fr fine-tuning.}
    \label{fig:size_effect_fr}
\end{figure}

\begin{figure}[h!]
    \centering
    \includegraphics[width=7.5cm]{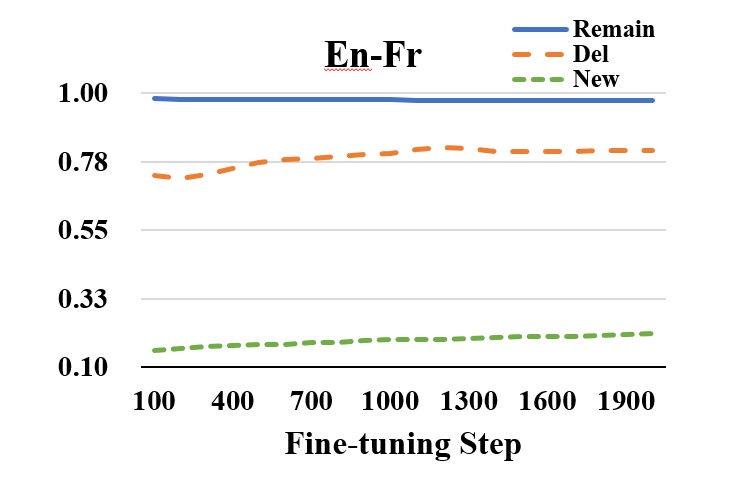}
    \caption{The self-correction effect in En-Fr fine-tuning process.}
    \label{fig:self-correction-process-enfr}
\end{figure}

The corpus size effect is presented in Figure \ref{fig:size_effect_fr}. It shows the trend same to Figure \ref{fig:datasize-for-finetune}, though the trend is not so significant for En-Fr. The self-correction effect is presented in Figure \ref{fig:self-correction-process-enfr}. The effect is the same to those in the other four language pairs. 

\begin{table}[ht]
\setlength\tabcolsep{12pt}
\small
\centering
\begin{tabular}{lcccc}
\hline
 &  \multicolumn{2}{c}{MaskAlign} & \multicolumn{2}{c}{Fine-Tuning mBERT}\\
\hline
& P & R & P & R\\
\hline
Zh-En & 81.9 & 87.3 & 88.8 & 88.1\\
De-En & 89.2 & 78.9 & 90.9 & 79.6 \\
En-Fr & 95.1 & 96.6 & 96.5 & 95.4 \\
Ro-En & 81.8 & 77.6 & 87.3 & 74.6 \\
Ja-En & 74.4 & 48.1 & 80.6 & 55.0\\
\hline
\end{tabular}
\caption{\label{tbl:PR-values}
Precision and Recall of MaskAlign predictions and the results of fine-tuning mBERT supervised by MaskAlign.
}
\end{table}

\section{Precision and Recall of Predicted Word Alignments}

Besides AER, we also evaluate the word alignment predictions by computing precision and recall using the gold alignments in the test sets. MaskAlign is used in this study due to its best performance among the third-party aligners. Its word alignments are used to supervise the fine-tuning of mBERT. The precision and recall are reported in Table \ref{tbl:PR-values}. It shows that precision is always significantly improved after the fine-tuning, while recall improvement is not so significant. On En-Fr and Ro-En, recall is slightly worse after the fine-tuning.

\section{40k Sentence Pairs for Both Training The Third-Party Aligner and Fine-tuning}
\begin{table}[h!]
\setlength\tabcolsep{2pt}
\small
\centering
\begin{tabular}{lcccccc}
\hline
 & Zh-En & De-En & En-Fr & Ro-En & Ja-En & AVG\\
\hline
w/o Fine-tuning & 17.9  &17.4  & 5.6  & 27.3 & 45.2 & 22.7  \\
MaskAlign$_{\rm adapted}$ & 82.2 & 36.4 & 19.9 & 51.1 & 90.6 & 55.9 \\
Fine-tuning & 17.3  &16.8  & 5.2  & 23.6 & 44.2 & 21.4  \\
\hline

\end{tabular}
\caption{\label{tbl:4w-maskalign}
AER of fine-tuning mBERT with the third-party supervision, which is generated by MaskAlign$_{\rm adapted}$ trained on 40k sentence pairs.
}
\end{table}

We use smaller parallel corpus, which consists of 40k sentence pairs for both training the third-party aligner and fine-tuning. Table~\ref{tbl:4w-maskalign} shows the result. AER of MaskAlign$_{\rm adapted}$ deteriorates sharply compared to training it on 80k sentence pairs shown in Table \ref{tbl:8w-maskalign}, but fine-tuning with such worse alignments as the supervision still gets better AER than without fine-tuning. We investigate the precision and recall of MaskAlign$_{\rm adapted}$ listed in Table \ref{tbl:PR-values-maskalign}, and find that it always obtains high precision, and these fewer but accurate alignments are useful supervision information for the fine-tuning.

\begin{table}[t!]
\setlength\tabcolsep{12pt}
\small
\begin{tabular}{lcccc}
\hline
 &  \multicolumn{2}{c}{80k} & \multicolumn{2}{c}{40k}\\
\hline
& P & R & P & R \\
\hline
Zh-En & 89.8 & 72.3 & 87.7 &9.9 \\
De-En & 92.9 & 62.0 & 94.5 &48.0  \\
En-Fr & 92.4 & 92.4 & 92.2 &67.3 \\
Ro-En & 85.0 & 70.4 & 87.3 &34.0 \\
Ja-En & 81.9 & 24.9 & 84.8 & 5.0 \\
\hline
\end{tabular}
\caption{\label{tbl:PR-values-maskalign}
Precision and Recall of MaskAlign$_{\rm adapted}$ predictions with different sizes of parallel training corpus .
}
\end{table}

\end{document}